%% file: root.tex
\documentclass[letterpaper, 10 pt, conference]{ieeeconf}  %

\IEEEoverridecommandlockouts                              %

\usepackage{graphicx}
\usepackage{amsmath}
\usepackage{bbm}
\usepackage{dsfont}
\usepackage{subcaption}

\PassOptionsToPackage{dvipsnames}{xcolor}
\PassOptionsToPackage{table}{xcolor}

\usepackage{amssymb}
\usepackage{booktabs}
\usepackage{makecell}
\usepackage[ruled,noend,linesnumbered]{algorithm2e}
\usepackage{titlesec}
\usepackage{wrapfig}
\usepackage{array}
\usepackage{multirow}
\usepackage{colortbl}
\usepackage{tabularx}
\usepackage{macros/mysymbols}

\usepackage{grffile}
\usepackage{xspace}
\usepackage{needspace}

\usepackage{comment} 
\usepackage{url}
\usepackage{bm}
\usepackage{paralist}

\usepackage{pifont}
\newcommand{\xmark}{\ding{55}}%

\usepackage[belowskip=0pt,aboveskip=5pt,font=small]{caption}
\usepackage[belowskip=0pt,aboveskip=0pt,font=small]{subcaption}
\usepackage{xcolor}
\PassOptionsToPackage{hyphens}{url}
\usepackage[hidelinks]{hyperref}
\usepackage{xfrac}
\input{macros/macros.tex}

\usepackage{cite}

\title{\LARGE \bf
    ViNL: Visual Navigation and Locomotion \emph{Over Obstacles}
}
\author{
  Simar Kareer$^{1*}$, Naoki Yokoyama$^{1*}$, Dhruv Batra$^{1}$, Sehoon Ha$^{1}$, and Joanne Truong$^{1}$\\
  \thanks{$^{1}$SK, NY, DB, SH, and JT are with Georgia Institute of Technology
{\tt\footnotesize \{skareer, nyokoyama, dbatra, sehoonha, truong.j\}@gatech.edu}}
}

\begin{document}
\bstctlcite{IEEEexample:BSTcontrol}

\twocolumn[{%
  \renewcommand\twocolumn[1][]{#1}%
  \maketitle
  \vspace{-0.5cm}
  \includegraphics[width=\textwidth]{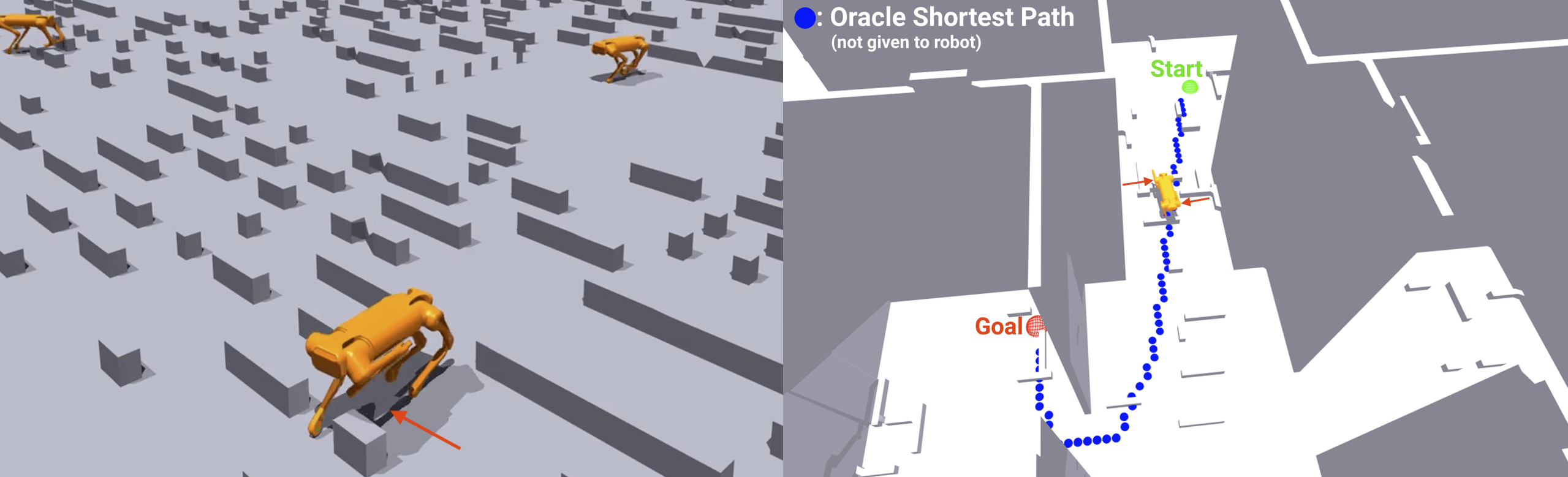}
  \label{fig:teaser}
  \vspace{-0.45cm}
  \captionof{figure}{\textbf{Left:} We learn a visual locomotion policy to follow linear and angular velocity commands while carefully avoiding obstacles with its feet using egocentric depth images. \textbf{Right:} We couple this locomotion policy with a high-level visual navigation policy, enabling AlienGo to navigate to goal locations in novel environments (without a map) while stepping over small obstacles on the ground.}
  \vspace{0.5cm}
}]

\let\svthefootnote\thefootnote
\let\thefootnote\relax\footnote{SK, NY, DB, SH, and JT are with the Georgia Institute of Technology {\tt\small\{skareer, nyokoyama, dbatra, sehoonha, truong.j\}@gatech.edu}}
\let\thefootnote\relax\footnote{\llap{\textsuperscript{*}}Denotes equal contribution.}
\addtocounter{footnote}{-1}\let\thefootnote\svthefootnote\

\vspace{-0.35cm}
\begin{abstract}
We present Visual Navigation and Locomotion over obstacles (ViNL), which enables a quadrupedal robot to navigate unseen apartments while stepping over small obstacles that lie in its path (e.g., shoes, toys, cables), similar to how humans and pets lift their feet over objects as they walk.
ViNL consists of: (1) a visual navigation policy that outputs linear and angular velocity commands that guides the robot to a goal coordinate in unfamiliar indoor environments; and (2) a visual locomotion policy that controls the robot’s joints to avoid stepping on obstacles while following provided velocity commands.
Both the policies are entirely `model-free', i.e. sensors-to-actions neural networks trained end-to-end. The two are trained independently in two entirely different simulators and then seamlessly co-deployed by feeding the velocity commands from the navigator to the locomotor, entirely `zero-shot' (without any co-training). 
While prior works have developed learning methods for visual navigation or visual locomotion, to the best of our knowledge, this is the first fully learned approach that leverages vision to accomplish both (1) intelligent navigation in new environments, and (2) intelligent visual locomotion that aims to traverse cluttered environments \textit{without} disrupting obstacles.
On the task of navigation to distant goals in unknown environments, ViNL using just egocentric vision \emph{significantly} outperforms prior work on robust locomotion using privileged terrain maps (+32.8\% success and -4.42 collisions per meter). 
Additionally, we ablate our locomotion policy to show that each aspect of our approach helps reduce obstacle collisions.  Videos and code at \href{http://www.joannetruong.com/projects/vinl.html}{http://www.joannetruong.com/projects/vinl.html}.

\end{abstract}

\input{sections/1_intro}
\input{sections/2_related_works}

\input{sections/3_method}
\input{sections/4_results}

\input{sections/5_conclusion}

\input{sections/6_acknowledgements.tex}

{
    \bibliographystyle{style/IEEEtran}
    \bibliography{bib/strings,bib/main}
}

\end{document}


\bstctlcite{IEEEexample:BSTcontrol}

\maketitle
\thispagestyle{empty}
\pagestyle{empty}

\vspace{-0.35cm}
\section{Visual Locomotion Policy Hyperparameters}
\xhdr{Reward}
\todo{write out reward, batch size, learning rate, training parameters}

\section{Visual Navigation Policy Hyperparameters}
\xhdr{Reward}
\todo{write out reward, architecture in math}
\cite{truong2020learning}
{
    \bibliographystyle{style/IEEEtran}
    \bibliography{bib/strings,bib/main}
}

%% file: macros/macros.tex
\newcommand{\xhdr}[1]{\vspace{2pt}\noindent\textbf{#1}}

\newcolumntype{Y}{>{\centering\arraybackslash}X}
\newcolumntype{P}[1]{\begin{center}>{\arraybackslash}p{#1}\end{center}}

%% file: sections/1_intro.tex
\section{Introduction}
For mobile robotic assistants to be useful in the real-world, they must skillfully navigate environments they have never seen.
This is particularly critical for indoor environments since they are subject to constant change (e.g., change in furniture layouts, temporary clutter, etc.).
In recent years, indoor navigation has seen significant progress using learned agents due to advances in deep reinforcement learning \cite{ddppo}, extensive datasets of real-world indoor scans \cite{ramakrishnan2021habitat}, and scalable photo-realistic simulation \cite{habitat19iccv, szot2021habitat, shen2021igibson, xiang2020sapien, deitke2020robothor}.
Works such as \cite{sct21iros, kadian2020sim2real, chaplot2020learning, Partsey_2022_CVPR, truong2022kin2dyn, robothor_challenge} show that agents trained entirely in simulation can be deployed in the real world in previously unseen environments without using a pre-computed map.

However, current progress in visual indoor navigation within previously unseen environments has been largely limited to using wheeled-base robots in homes with immaculate, flat terrain. Typical wheeled robots have limited maneuverability, which can pose a problem even in indoor home environments; they have difficulty going over clutter, doorways with thresholds, stairs, and thick carpets.
In contrast, legged robots are well-suited for navigating under such conditions; they can step over obstacles without disrupting or breaking them. 
While there are several works in learning legged locomotion, these works often use blind and reactive policies meant for outdoor environments, which emphasize stability and robust recovery on rough terrain \cite{rma, iscen2018policies}.

\input{figures/hierarchical_policy}

In our work we aim to bring legged robots to the unstructured and messy human world by enabling them to navigate \textit{over} clutter typically found in indoor environments.  We wish to replicate a cat's ability to carefully walk over small household obstacles (such as shoes, toys, clothes, etc.), rather than navigating around these obstacles entirely.
We present Visual Navigation and Locomotion over obstacles (ViNL), which enables a quadrupedal robot to navigate an apartment while stepping over small obstacles that lie in its path using only egocentric vision. 

Our low-level visual locomotion policy is learned in three stages using NVIDIA’s Isaac Gym simulation environment \cite{gym}. First, we learn to walk in a wide variety of challenging terrains using large-scale deep reinforcement learning (without any pre-training). 
Here the robot learns to walk and follow velocity commands with access to a privileged terrain map of height samples around the robot.
Next, we learn to walk over clutter by fine-tuning the previous policy in a novel terrain that contains small obstacles on the ground. In this stage, the robot learns to avoid stepping on the obstacles using the privileged terrain map. Finally, we learn to walk over clutter using egocentric vision using
supervised learning to reconstruct the privileged terrain map using vision alone. 
Unlike the other stages, the visual policy uses a LSTM \cite{hochreiter97lstm} to retain memory of its prior visual observations and proprioceptive states.  This enables the robot to walk over obstacles even when they leave the robot's view.

Separately, we train a high-level visual navigation policy in photo-realistic 3D scans of real-world indoor environments using the Habitat simulator \cite{habitat19iccv, szot2021habitat}. Similar to \cite{truong2022kin2dyn}, the navigation policy is trained using kinematic control in simulation, which commands robot center-of-mass linear and angular velocities. Although the locomotion and navigation policies are trained in two different simulators (Isaac Gym and Habitat), the two can be combined zero-shot for the task of visual navigation over obstacles. Our approach can navigate cluttered environments using only egocentric vision, up to 32.8\% more successfully than baselines that leverage privileged terrain maps, and up to 4.4 less collisions with obstacles per meter traveled.

The core contributions of our work are: 
\begin{asparaenum}
    \item We propose ViNL, the first approach to the best of our knowledge to accomplish both (1) intelligent navigation in new environments, and (2) intelligent visual locomotion to traverse cluttered environments \textit{without} disrupting obstacles. We show that ViNL can successfully guide the robot to the goal in cluttered indoor environments with a success rate of 73.6\%, a 32.8\% increase from prior works \cite{nvidia}.
    \item We show zero-shot sim2sim -- our visual navigation policy (trained in kinematic simulation with a low-level controller that simply teleports the robot) generalizes zero-shot to a different simulator (with rigid-body dynamics and a learned low-level locomotor).

\end{asparaenum}

%% file: figures/hierarchical_policy.tex
\begin{figure*}[t!]
\centering
    \vspace{0.07in}
    \includegraphics[width=1.0\textwidth]{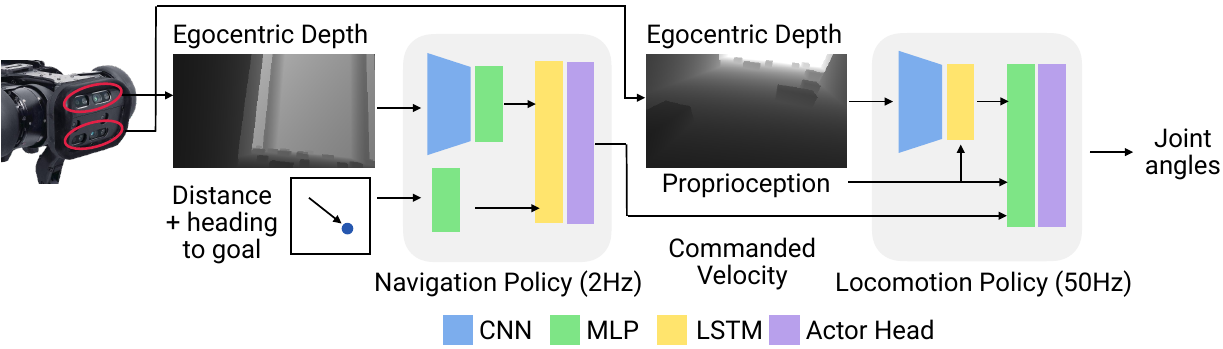}
    \caption{
        Visual Navigation and Locomotion over obstacles (ViNL) combines a high-level visual navigation policy with a locomotion policy to navigate over clutter in previously unseen indoor environments to a goal coordinate. The navigation policy runs at 2 Hz, while the locomotion policy runs at 50 Hz. The real-world AlienGo is equipped with two depth cameras. The navigation policy uses the top depth camera, pitched $15^{\circ}$ upwards, and the locomotion policy uses the bottom depth camera, pitched $30^{\circ}$ downwards.
    }
    \label{fig:hierarchical_policy}
\end{figure*}

%% file: sections/2_related_works.tex
\section{Related Work}
\xhdr{Legged Locomotion.}
Prior work in this area is significant and providing a comprehensive summary is beyond the scope of this manuscript. Most related is prior work on design of spanning the design of model-based controllers for a variety of different gaits \cite{raibert1990trotting, di2018dynamic, farshidian2017real, coros2010generalized, miura1984dynamic}, imitation learning for mimicking agile behaviors of animals \cite{peng2020learning}, and deep reinforcement learning approaches for directly learning locomotion controllers \cite{rma, nvidia, Energy, locomotion_terrain, miki2022learning}.
These locomotion works typically learn blind policies that demonstrate impressive stability while walking on rough terrain, slippery surfaces, slopes on real-world robots.
The primary focus of these blind locomotion works is on \textit{robustness and recovery}, whereas our work focuses on avoiding bad footholds and obstacle collisions altogether while navigating to a goal.

\xhdr{Visual Locomotion.} Some works have moved beyond blind locomotion and incorporated exteroceptive LiDAR sensors to obtain a map of the surrounding terrain before selecting a foothold to step on \cite{miki2022learning, Vladlen, locomotion_terrain, Deepgait}. 
However, LiDAR sensors that capture the terrain beneath the robot are expensive and absent from many legged platforms out-of-the-box (AlienGo, A1, and Go1 from Unitree, Spot from Boston Dynamics).

Recent works have investigated leveraging cameras instead as an inexpensive and more scalable solution. 
Several works use vision to guide foot placement over discontinuous terrains that feature gaps \cite{Deepgait, sehoonjie, margolis2021learning}. However, for indoor navigation, we argue that it is more important (and difficult) to step over \textit{obstacles}, rather than \textit{gaps}, which are more prevalent in daily environments. Stepping over obstacles requires understanding the relationship between the position of the robot's legs with respect to the height of the obstacle. In contrast, a policy that can step over gaps has not learned this, and may not raise its legs up high enough to step over obstacles. In concurrent work, \cite{agarwal2022legged} similarly leverage simulation to learn a policy to modulate the robot's gait to traverse challenging terrain such as stairs, and rocky terrain using a front-facing depth camera. However, this work focuses on visual locomotion, and does not couple the learned locomotion controller with navigation.

\xhdr{Visual Navigation.} 
Visual navigation is typically studied under static, social, or interactive environments \cite{anderson2018evaluation, gibson_challenge, xia2020interactive}, and has been shown to transfer to real-world environments for wheeled robots \cite{sct21iros, kadian2020sim2real, chaplot2020learning, Partsey_2022_CVPR}. In static navigation, a robot must navigate through an indoor environment that contains realistic arrangements of large furniture (tables, chairs, sofas, etc.). In social navigation, a robot must navigate to a goal location while avoiding collisions with pedestrians in the environment. Both static and social navigation ultimately results in the navigation policy learning to avoid large obstacles (furniture or moving pedestrians), and does not require fine-grain locomotion and control of legged robots, which is the focus of our work. In interactive navigation, a wheeled robot is able to push over small obstacles on its path. 
In our work, we aim to avoid interactions with the obstacles altogether, minimizing disturbances to the environment. 

Visual navigation has also been studied for legged robots using expert-designed \cite{truong2022kin2dyn, truong2020learning, sorokin2022learning} and fully learned \cite{jain2020pixels} low-level controllers. Additionally, \cite{couplingVision} leverages planning-based navigation methods with a learned locomotion policy to navigate in real-world environments. However, these works do not learn to use visual data to modulate the robot's gait to avoid obstacles. Again, the locomotion controllers are designed for robustness and recovery from collisions while following velocity commands, instead of preventing foot collisions altogether using vision.

%% file: sections/3_method.tex
\section{Method}

We propose ViNL, a hierarchical architecture (shown in Figure \ref{fig:hierarchical_policy}) for navigating over clutter in indoor environments, which consists of (1) a visual locomotion policy (shown in Figure \ref{fig:visual_locomotion}), trained to avoid stepping on small obstacles that operates at 50 Hz, and (2) a high-level visual navigation policy that commands linear and angular velocities at 2 Hz. The locomotion and navigation policies are trained in parallel, independently of each other.

\subsection{Visual Locomotion}
We utilize a three-stage approach for learning a low-level controller that walks over small obstacles. 

\xhdr{Stage 1: Learning to Walk.}
Before learning to walk over obstacles, we first aim to learn a teacher locomotion policy that can robustly walk over challenging terrain (\textit{e.g.}, stairs, rough terrain, etc.). 
Using the rough terrain environment in the Isaac Gym benchmark \cite{nvidia}, we teach the AlienGo quadruped robot using deep reinforcement learning (PPO) to walk over challenging terrain while following commanded linear and angular velocities.

The teacher locomotion policy takes as input the robot's proprioceptive state and a privileged height map $\bm{H}$ containing height samples from a 1.6m $\times$ 1.0m grid around the robot. Specifically, at timestep $t$, the observation space consists of:
joint positions $\bm{q}_{t} \in \mathbb{R}^{12}$,
joint velocities $\dot{\bm{q}_{t}} \in \mathbb{R}^{12}$,
previous joint commands $\bm{q}_{t-1} \in \mathbb{R}^{12}$,
base linear velocity \textbf{v} $\in \mathbb{R}^{3}$,
base angular velocity $\bm{\omega} \in \mathbb{R}^{3}$,
projected gravity vector $\bm{g} \in \mathbb{R}^{3}$,
commanded velocities ${(v_x, v_y, \omega_z)}^{*} \in \mathbb{R}^{3}$,
and a terrain map $\bm{H} \in \mathbb{R}^{11 \times 17}$.
The output of the locomotion policy consists of joint angles $\bm{q} \in \mathbb{R}^{12}$, that are used as target positions for PD motor controllers. We use the same rewards from \cite{nvidia}, which encourages the robot to follow the commanded velocities while maintaining a smooth, natural gait. The episode is terminated when the robot falls over. 

Both the actor and critic networks consist of a 3-layer MLP, where each layer has 256 hidden units. The actor network outputs a mean and standard deviation that parameterizes a diagonal Gaussian distribution, from which actions are sampled. Notably, in contrast to \cite{nvidia}, which concatenates the local map of the robot's terrain directly with the rest of the observations, we first encode the local map separately using a 3-layer MLP encoder $\phi$ with 128, 64, and 32 hidden units. This map encoding is then passed to the policy with the rest of the observations, as shown in Figure \ref{fig:visual_locomotion}. This allows us to use Learning by Cheating \cite{Cheating} in Stage 3 of our approach (described below), in which we distill the knowledge from the privileged map MLP encoder, into an encoder $\tau$ that uses egocentric depth observations instead. 

The policy is trained using Proximal Policy Optimization (PPO) \cite{ppo} in the Isaac Gym simulation environment \cite{gym} using 50 parallel environments and a batch size of 500. We train our policies on a single RTX 3070 GPU until convergence, which takes approximately 15 minutes.

\xhdr{Stage 2: Learning to Walk Over Clutter.} While the locomotion policy from Stage 1 is robust at walking, and can recover from collisions with uneven terrain, we now teach the robot to avoid stepping on obstacles altogether. 

\input{figures/obstacle_terrain}

We first construct an obstacle terrain consisting of small blocks on the ground, as shown in Figure \ref{fig:obstacleEnv}. The size of the blocks are randomized between 0.2m to 0.5m in width, 0.1m to 0.2m in height, and the length is kept fixed at 0.2m. The robots start in a patch of terrain without obstacles, and is given a randomly sampled linear and angular velocity to follow (between [0.0, 1.0] $\sfrac{m}{s}$ for linear velocity, and $\pm 60.0 \sfrac{\circ}{s}$ for angular velocity), forcing the robot to venture into the terrain cluttered with obstacles. 
We also utilize a curriculum of increasing obstacle density for training the robot to navigate over obstacles, similar to the game-inspired curriculum introduced in \cite{nvidia}.  Our terrain consists of 25, 8m $\times$ 8m tiles, in which each tile has a different density of obstacles. The sparsest tile contains 20 obstacles, and the densest tile contains 200 obstacles. 
Robots which are able to walk more than half the total tile distance in an episode are promoted to a more challenging tile in the terrain. Using this terrain and density curriculum, we fine-tune the locomotion policy from Stage 1 with an added penalty that discourages the robot from making contact with the obstacles on the ground with its feet. Specifically, the robot receives a penalty of $-k$ (we use $k=1$) if any leg is in contact with an obstacle in the environment.
\begin{equation}
    r_{contact}=
    \begin{cases} 
        -k & \text{if any leg in contact}\\
        0 & \text{otherwise}\\
    \end{cases}
\end{equation}

\input{figures/visual_locomotion}

We also trained a locomotion policy that can walk over clutter without the pre-training in Stage 1. This policy learns from scratch in our obstacle terrain using the full reward from \cite{nvidia}, plus our contact penalty. However, this policy resulted in unnatural gaits in which the robot would raise its foot up very high to avoid touching any obstacles. In contrast, by pre-training in the challenging terrain and teaching the robot how to walk first, the robot learns natural walking gaits before it attempts to avoid obstacles. Our results in Section \ref{sec:results} show that pre-training in challenging terrain and fine-tuning in the obstacle terrain results in a 19.2\% higher success rate than learning to walk over clutter directly.

\xhdr{Stage 3: Learning to Walk Over Clutter with Vision.} Finally, we aim to lift the assumption of access to a privileged terrain map for navigating over clutter. Instead of a terrain map $H$, we train the robot to avoid clutter using egocentric vision $d$ from a front-facing depth camera pitched downwards 30$^{\circ}$. Figure \ref{fig:timelapse} shows an example of our locomotion policy using vision to guide the robot to lift up its feet to avoid the obstacle in its path.

\input{figures/timelapse}

\looseness=-1
We use Learning by Cheating \cite{Cheating} to predict the encoding of a privileged terrain map $\phi(H)$ using an encoder that takes in egocentric depth observations and proprioceptive states of the robot while the policy is kept frozen. 
We use an LSTM $\tau$ provide the policy with temporal information. At timestep $t$, the LSTM $\tau$ takes in visual encodings of depth images from a CNN $\psi(d_t)$, proprioceptive states of the robot $p_t$, and its previous hidden state $h_{t-1}$ to predict the encoding of the current terrain map $\phi(H_t)$. 
\begin{equation}
\tau(\psi(d_t), p_t, h_{t-1}) \rightarrow \phi(H_t).
\end{equation}

An LSTM enables the policy to leverage temporal dependencies using the hidden state, without the costly compute and memory that is typically required for processing a large buffer of images. This is a key difference between our method and Kumar et al., which uses a history of 50 consecutive timesteps worth of previous observations. Leveraging temporal information is particularly important for navigating over clutter, as the robot only has access to a front-facing egocentric depth camera. As the robot walks over objects, and the objects leave the camera's view, it must make use of past temporal information in order to remember the location of these obstacles and avoid them with its legs. 

We train a 3-layer CNN $\psi$ and a 2-layer LSTM $\tau$ using supervised learning with on-policy data to minimize MSE loss while keeping the policy frozen, similar to \cite{rma, dagger}. Specifically, we use 30 parallel environments, and collect 10 steps of rollout for each, resulting in a batch size of 300. We train our policies on a single RTX 3070 GPU until convergence, which took approximately 4.5 hours. 
\looseness=-1
By using our multi-stage approach, we avoid rendering computationally expensive depth images when training our locomotion policies using deep reinforcement learning. Instead, we can learn robust locomotion polices very quickly (stages 1-2) using proprioception and terrain maps as observations, and only introduce depth images in the final stage during policy distillation. Using a single RTX 3070 GPU, we were able to simulate 30 robots each equipped with a depth camera at 190 FPS. In comparison, using the same compute, we can simulate 1024 robots sensing terrain maps at 23,000 FPS.

\subsection{Visual Navigation}
\xhdr{Task: PointGoal Navigation.} In PointGoal Navigation \cite{anderson2018evaluation}, a robot is tasked with navigating to a goal location  (typically 5-30m away) in an indoor environment without being given a pre-built map of the environment. The robot has access to an egocentric depth camera, as well as egomotion estimates that indicate its relative position and heading from its starting pose. The goal location is specified in polar coordinates, also relative to the agent's starting pose. An episode is considered successful when the robot reaches within 0.325m of the goal (half the length of the AlienGo robot we use).

\looseness=-1
\xhdr{Dataset.} We use both the Habitat-Matterport (HM3D) \cite{ramakrishnan2021habitat} and Gibson \cite{xia2018gibson} 3D datasets, which contains over 1000 scans of real-world indoor environments (homes, offices, etc.). Examples of scans used for training are shown in Figure \ref{fig:hm3d_train}. Notice the diversity, realism, and size (multiple rooms) of the environments.

\input{figures/hm3d_visualization}

\input{figures/main_table}

\xhdr{Kinematic Visual Navigation.} The navigation policy takes as input an egocentric depth image and the robot's current distance and heading relative to the goal.
The latter can be computed from the egomotion estimate and the goal coordinate, as both are specified relative to the robot's starting pose.
The output of the navigation policy is a 2-dimensional vector representing the desired center-of-mass forward linear and angular velocities ($v_x, \omega_z$), where $v_x$ and $\omega_z$ have ranges of [0.0, 1.0] $\sfrac{m}{s}$ and $\pm 60.0 \sfrac{^\circ}{s}$, respectively.
A ResNet-18 \cite{he2016resnet} visual encoder is used to process the egocentric depth images. The goal vector is encoded using a single linear layer, whose output is concatenated to the output of the visual encoder and then passed into a downstream 2-layer LSTM policy. Similar to the locomotion policy, the final layer of the policy outputs a mean and standard deviation that parameterizes a diagonal Gaussian distribution from which actions are sampled. The overall architecture of the navigation policy is shown in Figure \ref{fig:hierarchical_policy}. We use the same reward function from \cite{truong2022kin2dyn}, which encourages path efficiency and discourages collisions with the environment. 

For training the navigation policy, we use kinematic control as an approximation for the robot's movement. At each step, we directly teleport the robot (without running full rigid-body dynamics simulation) by integrating the commanded velocities at 2 Hz. If the resulting pose intersects with the environment, the robot is simply kept in its current pose until valid velocity commands are given. This formulation was shown in \cite{truong2022kin2dyn} to lead to better sim-to-real transfer by abstracting away low-level physics interactions between the robot and its environment, in favor of gathering more experience through faster simulation.

However, since the robot is controlled kinematically during navigation training, the depth observations from the robot are completely level with the ground, which does not accurately reflect the camera shake that occurs when mounted on moving quadruped robots. To improve the robustness of our navigation policies against camera shake, we randomize the roll and pitch of the front-facing camera by up to $\pm 30^{\circ}$ in either direction at each step during training.

\looseness=-1
The policy is trained using DD-PPO \cite{ddppo}, a distributed reinforcement learning method, using the Habitat simulator \cite{habitat19iccv, szot2021habitat}. Habitat supports fast rendering of photo-realistic 3D simulation environments, which prior work has found to successfully transfer to both wheeled and legged robots in the real-world \cite{sct21iros, kadian2020sim2real, truong2020learning}. We train our policy using 8 GPUs for 3 days (500M steps of experience) to ensure convergence.

%% file: figures/obstacle_terrain.tex
\begin{figure}[ht]
    \centerline{\includegraphics[width=1.0\columnwidth]{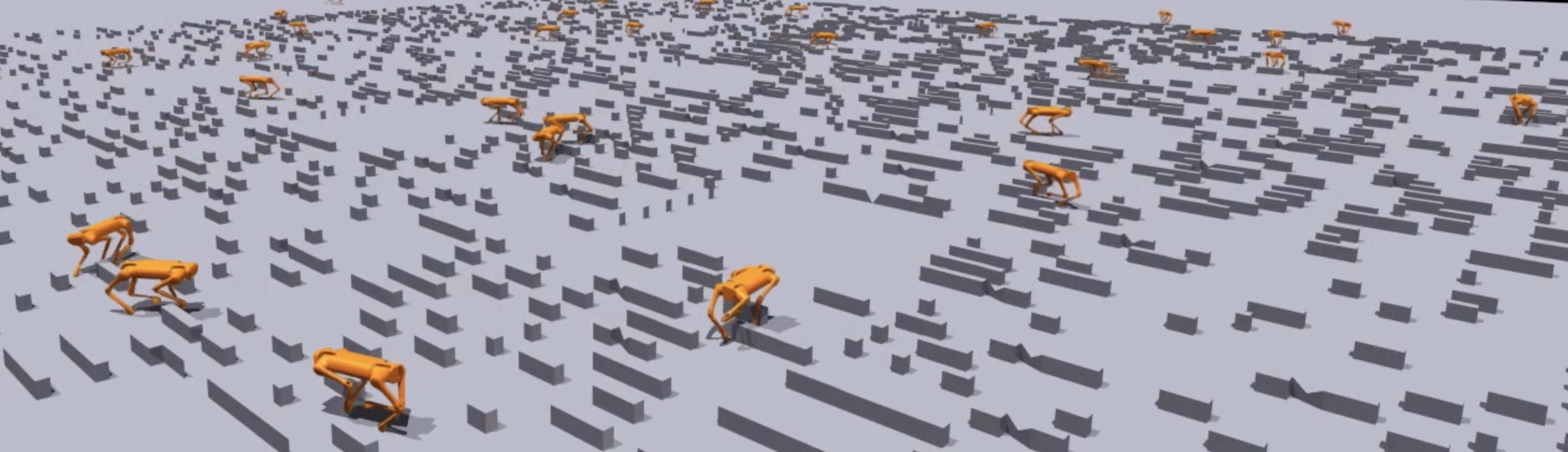}}%
    \caption{
        The obstacle terrain is cluttered with small blocks.
    }
    \vspace{-0.4cm}
    \label{fig:obstacleEnv}
\end{figure}

%% file: figures/visual_locomotion.tex
\begin{figure}[t!]
    \vspace{0.08in}
    \includegraphics[width=1.0\columnwidth]{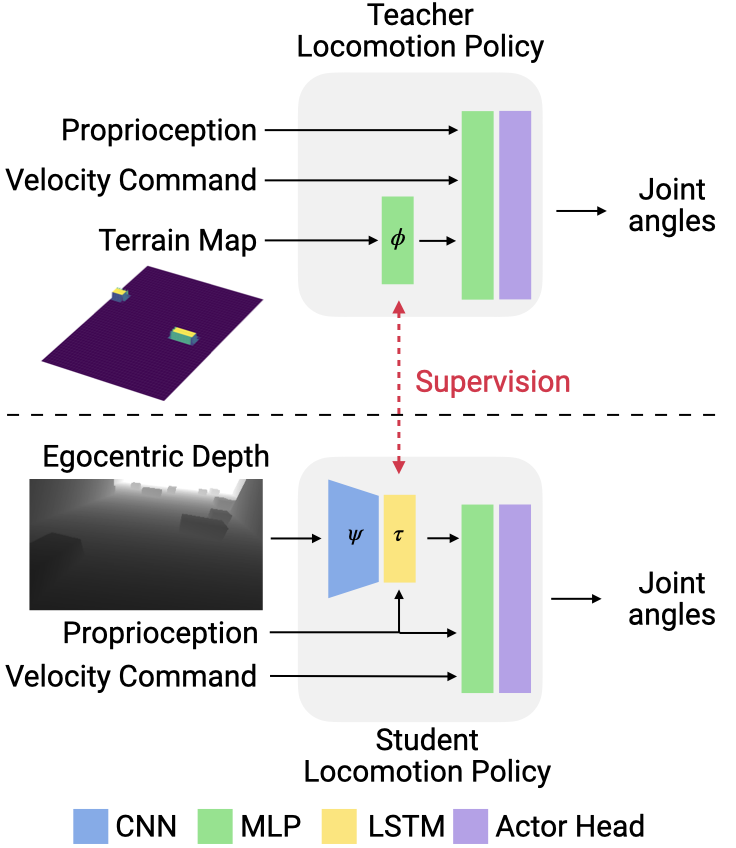}
    \caption{
        We train a visual locomotion policy to step over obstacles using egocentric vision. We first learn a teacher locomotion policy (top) that has access to privileged terrain map of the environment, and then distill this information into a student locomotion policy (bottom) that uses egocentric depth observations. 
    }
    \vspace{-.3cm}
    \label{fig:visual_locomotion}
\end{figure}

%% file: figures/timelapse.tex
\begin{figure}[h]
    \vspace{0.07in}
    \centerline{\includegraphics[width=1.0\columnwidth]{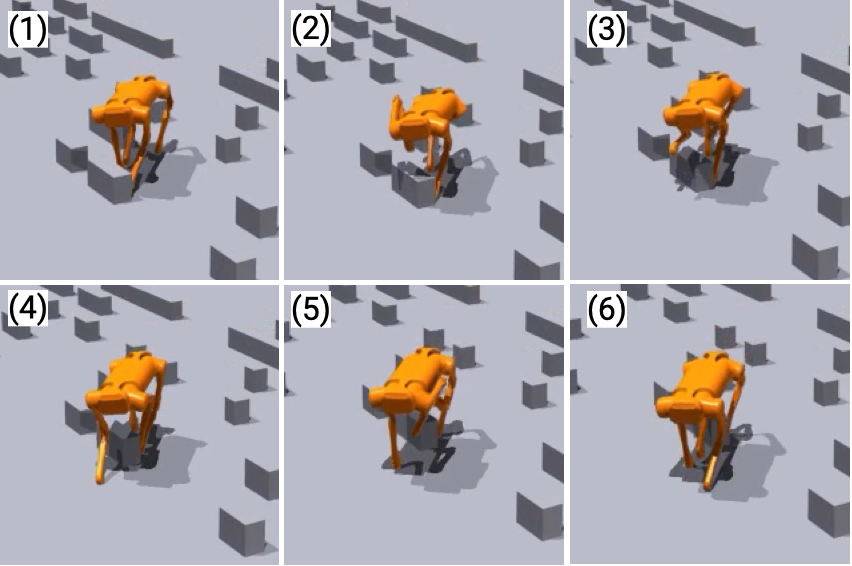}}
    \caption{
        A time-lapse demonstrating our locomotion policy navigating over obstacles. The robot first approaches the obstacle (frame 1), lifts one leg over the block (frames 2, 3), and places the leg over the obstacle (frame 4). The robot repeats this motion for its other legs, successfully walking \textit{over} the obstacle in its path.
    }
    \vspace{-0.2cm}
    \label{fig:timelapse}
\end{figure}

%% file: figures/hm3d_visualization.tex
\begin{figure}[h]
    \centerline{\includegraphics[width=1.0\columnwidth]{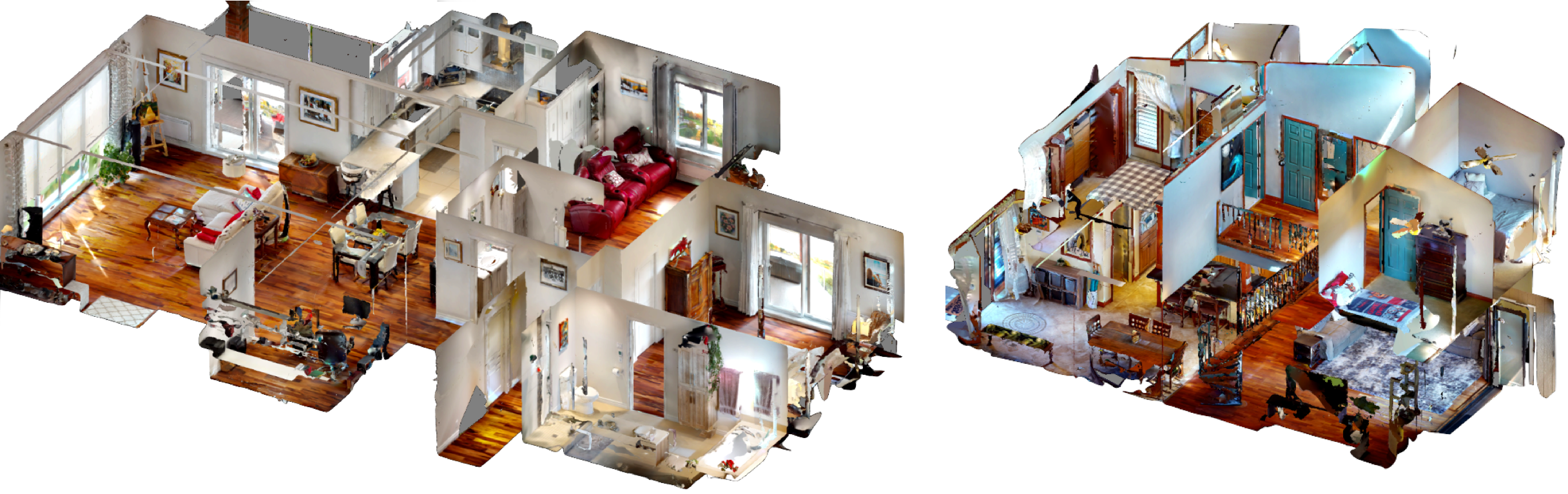}}
    \caption{
        Examples of real-world scans from the HM3D dataset. %
    }
    \label{fig:hm3d_train}
    \vspace{-0.2cm}
\end{figure}

%% file: figures/main_table.tex
\begin{table*}[t!]
\vspace{0.06in}
\centering
\scriptsize
\caption{\small We evaluate all policies in cluttered floor plans simulated in Isaac Sim. 
Our method is both more successful and better at avoiding collisions with obstacles than all other baselines.}

\resizebox{\textwidth}{!}{
\label{tab:results}
    \begin{tabular}{cccccccccc}
        \toprule
        \multirow{2}{*}{\textbf{\#}}  & \multirow{2}{*}{\textbf{Method}} & \multicolumn{2}{c}{\bf{Sensors}} & \multicolumn{2}{c}{\bf{Train Terrain}} & \multirow{2}{*}{\shortstack{\textbf{Leg Contact}\\ \textbf{Penalty}}} &
        \multirow{2}{*}{\shortstack{\textbf{Success}\\ \textbf{Rate $\uparrow$}}} &
        \multirow{2}{*}{\shortstack{\textbf{Distance}\\ \textbf{traveled (m) $\uparrow$}}} &
        \multirow{2}{*}{\shortstack{\textbf{Foot Collisions}\\ \textbf{per Meter $\downarrow$}}} \\
        \cmidrule(l{4pt}r{4pt}){3-4} \cmidrule(l{4pt}r{4pt}){5-6}
        & & \textbf{Terrain Map} & \textbf{Vision} & \textbf{Rough} & \textbf{Obstacle} & & & & \\
        \midrule
        1 & Blind & \xmark & \xmark & \checkmark & \checkmark & \checkmark & 1.20 $\pm$ 0.98 & \textbf{10.20 $\pm$ 0.83} & 18.23 $\pm$ 5.31 \\
        2 & Rough & \checkmark & \xmark & \checkmark & \xmark & \xmark & 40.80 $\pm$ 3.49 & 7.59 $\pm$ 0.17 & 16.50 $\pm$ 2.50 \\
        3 & ViNL (no-pretraining) & \xmark & \checkmark & \xmark & \checkmark & \checkmark & 54.40 $\pm$ 7.94 & 6.55 $\pm$ 0.98 & 14.09 $\pm$ 7.05 \\
        4 & ViNL (MLP) & \xmark & \checkmark & \checkmark & \checkmark & \checkmark & 66.80 $\pm$ 8.06 & 7.54 $\pm$ 0.69 & 12.16 $\pm$ 0.35 \\
        5 & ViNL & \xmark & \checkmark & \checkmark & \checkmark & \checkmark &  \textbf{73.60 $\pm$ 5.85} & 8.19 $\pm$ 0.34 & \textbf{12.08 $\pm$ 0.72} \\
        \midrule
        \multirow{2}{*}{6} & \multirow{2}{*}{\shortstack{ViNL \\ Eval. w/ no obstacles}}  & \multirow{2}{*}{\xmark} & \multirow{2}{*}{\checkmark} & \multirow{2}{*}{\checkmark} & \multirow{2}{*}{\checkmark} & \multirow{2}{*}{\checkmark} & \multirow{2}{*}{86.80 $\pm$ 4.66} & \multirow{2}{*}{7.67 $\pm$ 0.35} & \multirow{2}{*}{2.90 $\pm$ 0.32} \\
        \\
        \bottomrule
    \end{tabular}
    }
    \vspace{-0.15in}
\end{table*}

%% file: sections/4_results.tex
\section{Results}
\label{sec:results}
In this section, we aim to address the following questions:
\begin{enumerate}
    \item How well can our visual navigation policy perform in a different simulator and different low-level controller?
    \item How well does ViNL navigate while stepping over obstacles compared to prior work?
    \item How vital is the use of exteroception, pre-training, or LSTM layers for the locomotion policy?
\end{enumerate}

We evaluate using the task of PointGoal Navigation over clutter across 5 unique floor plans in Isaac Gym as shown on the right of Figure 1. In each floor plan, we randomly sample 10 start and goal locations for the robot, on average 7m apart. The environments mimic rooms laid out in a a maze-like pattern and navigation can require exploration and backtracking.  We report results across 5 seeds for a total of 250 experiments per method.
During evaluation, the navigation policy uses a front-facing depth camera pitched $15^{\circ}$ upwards, while the locomotion policy uses a front-facing depth camera pitched $30^{\circ}$ downwards, as shown in Figure \ref{fig:hierarchical_policy}.

\looseness=-1
To get an upper bound for the performance of ViNL, we first evaluate the performance of our visual navigation policy for PointGoal Navigation without clutter. 
This allows us to compare the performance of the navigator `in-domain' (Habitat w/ kinematic low-level control) vs `out-of-domain' (Isaac Gym w/ low-level control using ViNL).
In-domain, the visual navigation policy is evaluated using 1000 episodes from scenes in the validation split of the HM3D dataset, and achieves a success rate (SR) of 89.70\%. Out-of-domain, ViNL has a SR of 86.8\% (Table \ref{tab:results}, row 6), indicating a fairly small sim-to-sim gap (2.9\% SR). This demonstrates that despite training the high-level navigation policy and low-level locomotion policy \textit{separately in different simulators and different levels of abstraction}, the two can be nearly seamlessly coupled together with a minimal drop in navigation performance. Since ViNL is evaluated in environments without clutter, the 2.9 foot collisions per meter reported in Table \ref{tab:results}, row 6 are from collisions against the walls in the environment.

Next, we compare ViNL to other baselines and ablations. In our experiments, we couple the same high-level navigation policy with the following locomotion policies:

\begin{asparaenum}
    \item \textbf{Blind:} Locomotion policy trained in rough terrains, and fine-tuned in the obstacle terrain with the leg contact penalty. This policy only has access to proprioception. This baseline measures the value of proprioception vs extroceptive sensing.
    \item \textbf{Rough:} Locomotion policy from \cite{nvidia} trained from scratch in rough terrains, without the leg contact penalty. This policy has access to a privileged terrain map for locomotion. This comparison establishes the value of using vision to step over obstacles vs privileged terrain maps for robustness and recovery for locomotion in complex terrain.
    \item \textbf{ViNL (no pre-training):} Same as ViNL, but all training is done in the obstacle terrain
    This locomotion policy uses egocentric depth observations. This experiment quantifies the importance of our pre-training phase for learning natural gaits and improving the robustness.
    \item \textbf{ViNL (MLP):} Same as ViNL, but the encoder is an MLP instead of an LSTM. This locomotion policy uses egocentric depth observations. This experiment measures the value of temporal information (MLP vs. LSTM).
\end{asparaenum}

All baselines are trained for the same amount of experience, including steps used for fine-tuning (500M steps). Note that the navigation policy is being tested out-of-distribution using a different simulator, novel environments, and unseen low-level controllers with no adaptation. 

\looseness=-1
For evaluation, we report success rate (SR), distance traveled, and foot collisions per meter. The foot collisions per meter are calculated across episodes in which the robot has navigated for at least 1m. This is to ensure that the foot collisions are not artificially lowered by short episodes in which the robot falls over close to the start. Additionally, since the low-level controller is being run at 50 Hz, a foot contact lasting 1 second will count as 50 foot collisions. We report the mean and standard deviation for all metrics in Table \ref{tab:results}.

From Table \ref{tab:results}, we see that \textbf{Blind} completely struggles (1.2\% SR, row 1), demonstrating the importance of using exteroceptive sensors in cluttered environments. Because the robots are spawned in obstacle-free patches of terrain during training and evaluation, we find that \textbf{Blind} learns to crawl in circles near the starting position and largely ignore commanded velocities in favor of not crashing due to obstacle collision. This results in a higher distance traveled, despite a near-zero success rate, and a foot collisions per meter that is not as large as one may expect from a blind policy, but it is ultimately an uncontrollable locomotor.

In comparison to \textbf{Rough}, we see that our method results in an average increase of 32.8\% SR, and average decrease of 4.42 foot collisions per meter traveled (rows 2 and 5), despite the fact that \textbf{Rough} has access to a privileged terrain map. We find that \textbf{Rough} often gets its hind feet stuck while climbing over an obstacle, causing the robot to fall over.

Next, we compare against \textbf{ViNL (no-pretraining)}, which was trained solely in the obstacle terrain. \textbf{ViNL (no-pretraining)} achieves a slightly higher success rate than \textbf{Rough} (+13.6\% SR, rows 2 and 3), and fewer foot collisions per meter (-2.41 collisions). However, this method still has a lower success rate than \textbf{ViNL} (-19.2\% SR, rows 3 and 5), and more foot collisions per meter (+2.01 collisions). This emphasizes the importance of pre-training in the rough terrain in stage 1 of our approach to avoid falling over.

We see that \textbf{ViNL (MLP)}, which was pre-trained on rough terrain, fine-tuned on obstacle terrain, and uses the leg contact penalty, performs just as well as \textbf{ViNL} for foot collision per meter (12.6 vs. 12.08, rows 4 and 5).
However, \textbf{ViNL} outperforms \textbf{ViNL (MLP)} by +6.8\% SR, which highlights the benefit of leveraging temporal information with an LSTM. Because \textbf{ViNL (MLP)} does not have any information about the whereabouts of objects that have left the robot's field-of-view, it cannot avoid obstacles that it tries to step over as adequately. This causes the robot to trip and fall over, leading to more episode failures.

%% file: sections/5_conclusion.tex
\section{Conclusion}
We present an approach that enables a robot to effectively navigate cluttered indoor environments using egocentric vision. While prior works focus on recovering from instability, or path planning around obstacles, we present a locomotion policy that avoids these obstacles using fine-grained control of its legs. Our work presents a fully learned hierarchical approach for both navigation and locomotion. These modules are trained separately but integrated seamlessly, without additional fine-tuning.

%% file: sections/6_acknowledgements.tex
\section{Acknowledgements}
\scriptsize{
We would like to thank Marco Delgado for help with real-world experiments using AlienGo. The Georgia Tech effort was supported in part by NSF, ONR YIPs, ARO PECASE, and Korea Evaluation Institute of Industrial Technology (KEIT) funded by the Korea Government (MOTIE) under Grant No.20018216, Development of mobile intelligence SW for autonomous navigation of legged robots in dynamic and atypical environments for real application. JT was supported by an Apple Scholars in AI/ML PhD Fellowship. The views and conclusions are those of the authors and should not be interpreted as representing the U.S. Government, or any sponsor. 

}